\documentclass[a4paper, fleqn]{cas-sc}

\usepackage[square, numbers, sort&compress]{natbib}
\usepackage[figurename=Fig.]{caption}
\usepackage{subfig}
\usepackage{graphicx}
\usepackage{mathrsfs}
\usepackage[scr=esstix, cal=boondox]{mathalfa} 
\usepackage{tikz}
\usepackage[capitalise]{cleveref}
\usepackage{array, multirow}
\usepackage{makecell}
\newcommand{\HighlightColor}{black}

\ExplSyntaxOn 
\keys_set:nn
{ stm / mktitle }
{ nologo } 
\ExplSyntaxOff 

\def\tsc#1{\csdef{#1}{\textsc{\lowercase{#1}}\xspace}}
\tsc{WGM}
\tsc{QE}
\tsc{EP}
\tsc{PMS}
\tsc{BEC}
\tsc{DE}

\begin{document}
  \let\WriteBookmarks\relax \def\floatpagepagefraction{1} \def\textpagefraction{.001}

  \shorttitle{GINOT}

  \shortauthors{Q. Liu et~al.}

  \title[mode = title]{Geometry-Informed Neural Operator Transformer}


  \cortext[corresponding]{Corresponding author}
  \author%
  [ncsa,ksu]{Qibang Liu}[orcid=0000-0001-7935-7907] \cormark[1] \ead{qibang@illinois.edu}

  \author%
  [cee]{Weiheng Zhong}
  \author%
  [cee]{Hadi Meidani}

  \author%
  [ncsa, nyu]{Diab Abueidda}

  \author%
  [ncsa,mse]{Seid Koric}

  \author%
  [bi,ae]{Philippe Geubelle}

  \affiliation[ncsa]{organization={National Center for Supercomputing Applications, University of Illinois Urbana-Champaign}, city={Urbana}, postcode={61801}, state={IL}, country={USA}}
  \affiliation[bi]{organization={Beckman Institute for Advanced Science and Technology, University of Illinois Urbana-Champaign}, city={Urbana}, postcode={61801}, state={IL}, country={USA}}
  \affiliation[ae]{organization={The Grainger College of Engineering, Department of Aerospace Engineering, University of Illinois Urbana-Champaign}, city={Urbana}, postcode={61801}, state={IL}, country={USA}}
  \affiliation[mse]{organization={The Grainger College of Engineering, Department of Mechanical Science and Engineering, University of Illinois Urbana-Champaign}, city={Urbana}, postcode={61801}, state={IL}, country={USA}}
  \affiliation[cee]{organization={The Grainger College of Engineering, Department of Civil and Environmental Engineering, University of Illinois Urbana-Champaign}, city={Urbana}, postcode={61801}, state={IL}, country={USA}}
  \affiliation[nyu]{organization={Civil and Urban Engineering Department, New York University Abu Dhabi}, country={United Arab Emirates}}
  \affiliation[ksu]{organization={Department of Industrial and Manufacturing Systems Engineering, Kansas State University}, city={Manhattan}, postcode={66506}, state={KS}, country={USA}}

  \begin{abstract}
    Machine-learning-based surrogate models offer significant computational efficiency
    and faster simulations compared to traditional numerical methods, especially
    for problems requiring repeated evaluations of partial differential
    equations. This work introduces the Geometry-Informed Neural Operator Transformer
    (GINOT), which integrates the transformer architecture with the neural
    operator framework to enable forward predictions on arbitrary geometries. GINOT
    employs a sampling and grouping strategy together with an attention mechanism
    to encode surface point clouds that are unordered, exhibit non-uniform point
    densities, and contain varying numbers of points for different geometries.
    The geometry information is seamlessly integrated with query points in the
    solution decoder through the attention mechanism. The performance of GINOT is
    validated on multiple challenging datasets, showcasing its accuracy and
    generalization capabilities for complex and arbitrary 2D and 3D geometries.
  \end{abstract}

  \begin{graphicalabstract}
    \includegraphics[width=\textwidth]{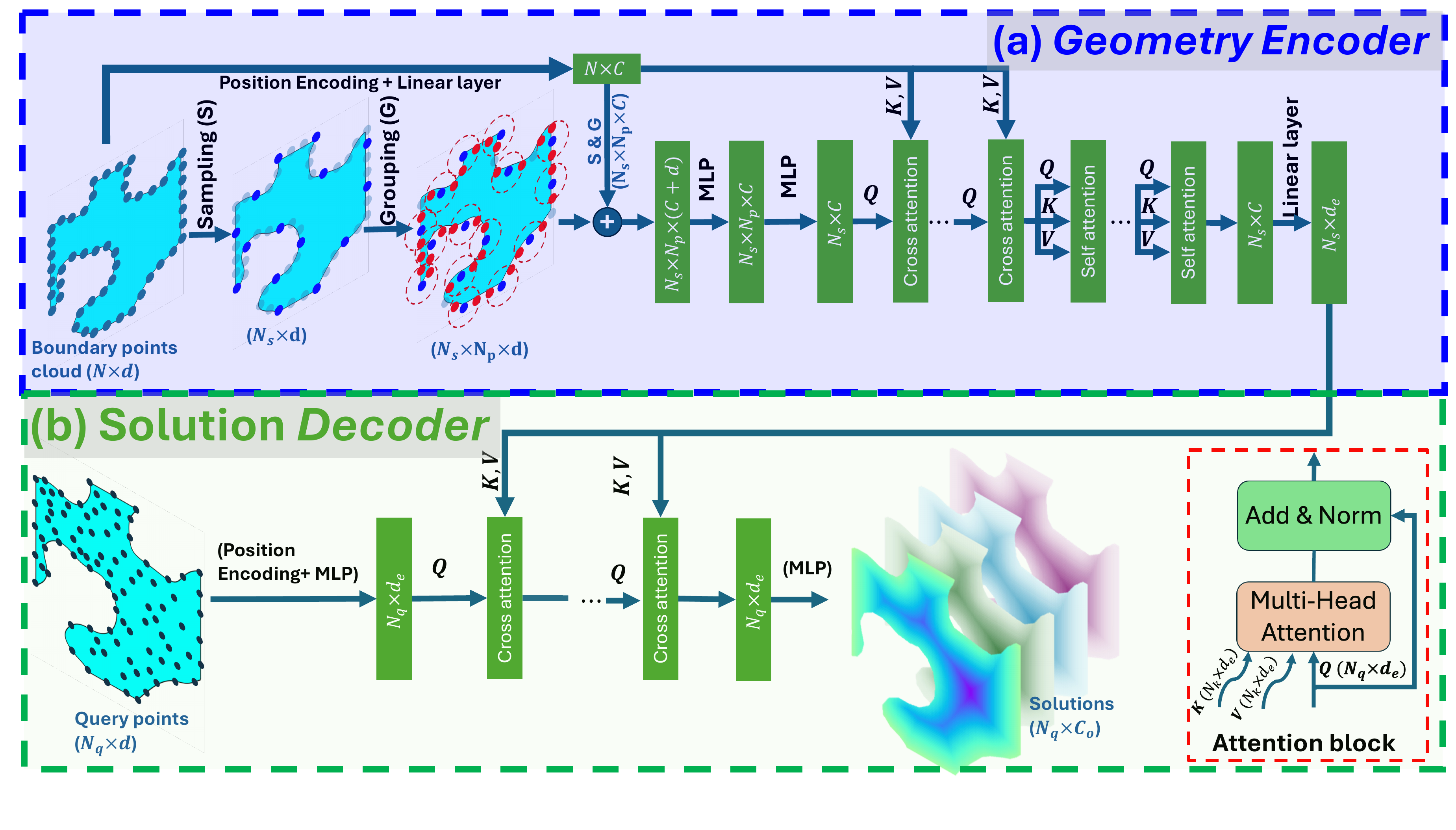}
  \end{graphicalabstract}

  \begin{highlights}
    \item A Geometry-Informed Neural Operator Transformer (GINOT) is proposed for
    forward predictions on arbitrary geometries

    \item GINOT encodes surface point clouds that are unordered, have non-uniform
    point density, and varying numbers of points.

    \item GINOT effectively processes complex, arbitrary geometries and varying input
    conditions with good predictive accuracy
  \end{highlights}

  \begin{keywords}
    Transformer \sep Geometry-informed \sep Neural operator \sep Arbitrary geometry
    \sep Point cloud \sep Deep learning
  \end{keywords}

  \maketitle

  \section{Introduction}

  Many natural phenomena in the physical world are governed by partial
  differential equations (PDEs), which encapsulate fundamental principles of physics.
  Numerical techniques such as finite difference, finite element, and finite
  volume methods have been developed to solve these equations computationally. While
  these methods enable high-resolution simulations to capture complex physical behaviors,
  they often demand significant computational resources, making them time-intensive,
  costly, and sometimes infeasible under current computational limitations.

  {\color{\HighlightColor}%
  Data-driven surrogate models have emerged to alleviate these computational burdens. Leveraging deep neural architectures, they approximate PDE solution operators with substantial speedups while retaining acceptable accuracy, making them effective when many evaluations are required \citep{liu2024adaptive}. Typical downstream tasks include sensitivity analysis \citep{kihara2019estimating,ankenbrand2021sensitivity}, gradient-based optimization \citep{dogo2018comparative,daoud2023gradient}, uncertainty quantification \citep{abdar2021review,caldeira2020deeply,Zhong2023PIVAE}, digital twins \citep{chattopadhyay2024oceannet,kobayashi2024deep,kobayashi2024improved}, and inverse or generative design \citep{liu2025univariate,liu2025towards}. These surrogates do not replace a single high-fidelity simulation; rather, they amortize its upfront cost over large query workloads, enabling dramatically faster iterative engineering and design cycles. %
  } %

  Neural operators have become prominent surrogate models, adept at mapping
  infinite-dimensional input functions to output functions at specified query
  points:
  \begin{equation}
    \label{eq:NOT}G_{\theta}:\mathcal{F}\rightarrow \mathcal{G}(\textbf{x}),
  \end{equation}
  where ${G}_{\theta}$ is the neural operator with learnable parameter $\theta$,
  $\mathcal{F}$ is the space of input functions, and $\mathcal{G}$ is the space of
  output functions on query points $\textbf{x}$. Examples such as the Fourier
  Neural Operator (FNO) \citep{li2020fourier} and the Deep Operator Network (DeepONet)
  \citep{lu2021learning} have shown significant success in this area \citep{kovachki2021universal,you2022learning,bonev2023spherical,cai2024towards,he2023novel,li2023phase,he2024sequential,gano,dcon, fc4no}.
  These models map inputs like initial and boundary conditions, material properties,
  and other functions to solution fields at query points. However, their
  applicability is often limited to simple geometries. For instance, FNO is
  tailored for fixed, regular meshes such as image grids or 3D voxels and requires
  a complete mesh graph as input due to its reliance on the Fast Fourier Transform
  kernel, making it unsuitable for arbitrary geometries and less scalable. While
  DeepONet can predict solution fields at arbitrary query points, its original
  formulation is restricted to single geometries with fixed query point
  locations and counts.

  However, real-world problems frequently involve complex and arbitrary computational
  geometries. Recent advances, including graph neural operators (GNO) \citep{li2020neural},
  Geo-FNO \citep{li2023fourier}, GI-FNO \citep{li2023GIFNO}, and Geom-DeepONet \citep{he2024geom},
  have addressed challenges in handling diverse geometries. GNO can process
  irregular grids using graph representations but is limited to local operations
  under constrained computational budgets, resulting in lower accuracy. Both Geo-FNO
  and GI-FNO rely on projecting irregular meshes onto fixed regular grids and
  mapping them back. However, Geo-FNO requires a fixed number of query points
  during training, while GI-FNO constructs new graphs on regular meshes using local
  kernel integration layers through GNO, which need longer inference time. Geom-DeepONet,
  while overcoming the limitation of single geometries of DeepONet by removing
  or repeating nodes to ensure fixed data size, is restricted to parametric geometries
  (e.g., length, thickness, radius) and does not generalize to fully arbitrary shapes.
  Additionally, GI-FNO requires a signed distance function (SDF) as input to represent
  geometry, and Geom-DeepONet depends on both geometry parameters and the SDF. However,
  obtaining the SDF is not always easily, particularly for complex 3D geometries
  represented by surface meshes, as it incurs a computational cost of
  $O(N_{g}\times N_{m})$ for distance calculation and $O(N_{g}\times N_{m})$ for
  the minimum value, where $N_{g}$ is the number of grid points in the domain and
  $N_{m}$ is the number of surface mesh—both of which are large for intricate geometries.

  Researchers have leveraged the attention mechanism of transformers \citep{Ashish2017atten}
  in neural operators to efficiently integrate information from input functions
  and query points \citep{cao2021choose,liu2022ht,li2022transformer,hao2023gnot}.
  Neural Operator Transformers (NOT) are particularly advantageous for irregular
  meshes due to their attention mechanism, enabling the prediction of solution fields
  at arbitrary query points. Unlike DeepONet, which employs a simple dot product
  to combine information from input functions and query points, the attention
  mechanism in NOT directs each query point to focus on the most relevant
  information from the input functions. This capability allows NOT to generalize
  effectively and capture more complex relationships between input functions and
  query points. Building on this, the authors developed an SDF-based NOT that accommodates
  arbitrary geometries \citep{liu2025towards}. The SDF representation facilitates
  handling arbitrary geometries on fixed, regular grids, which are compatible with
  existing deep learning models. However, as noted earlier, obtaining the SDF is
  not always straightforward or computationally easily. %

  In this work, we introduce the Geometry-Informed Neural Operator Transformer (GINOT),
  which integrates the transformer architecture with the neural operator framework
  to enable forward predictions for arbitrary geometries without requiring SDF as
  geometry features. GINOT consists of two primary components: the geometry encoder
  and the solution decoder. Drawing inspiration from Shap-E \citep{jun2023shap} by
  OpenAI, we design the geometry encoder to process the boundary point cloud, representing
  the geometry, into the KEY and VALUE matrices for the attention mechanism. This
  choice facilitates the integration of information from query points in the solution
  decoder.

  The geometry encoder utilizes sampling and grouping layers \citep{qi2017pointnet,qi2017pointnet++}
  to extract local geometric features from the boundary point cloud. These local
  features are fused with global geometry information through cross-attention layers,
  enabling effective processing of unordered and non-uniformly distributed point
  clouds—unlike convolutional neural networks (CNNs), which operate on regular grids
  with uniform density. Because the number of points in the point cloud can vary
  between geometries, padding is applied to standardize input sizes within a batch.
  To ensure that padding points do not influence the geometry encoder's output, a
  masking mechanism is employed: it prevents padding points from being selected
  in the sampling and grouping layers and excludes them from contributing to
  attention scores in the cross-attention layer.

  In addition to varying geometries, many problems involve additional input parameters
  such as loading, material properties, and boundary conditions. To address such
  cases, we propose an extension strategy for GINOT by incorporating additional encoders
  to process these input functions.

  This manuscript summarizes the key concepts, methodologies, and results of the
  GINOT model. We validate the effectiveness of GINOT on multiple challenging
  datasets and compare its performance with baseline methods. The results demonstrate
  that GINOT achieves high accuracy and generalizes effectively to arbitrary geometries.

  \section{GINOT}
  \label{sec:ginot}

\begin{figure}[h]
    \centering
    \includegraphics[width=\textwidth]{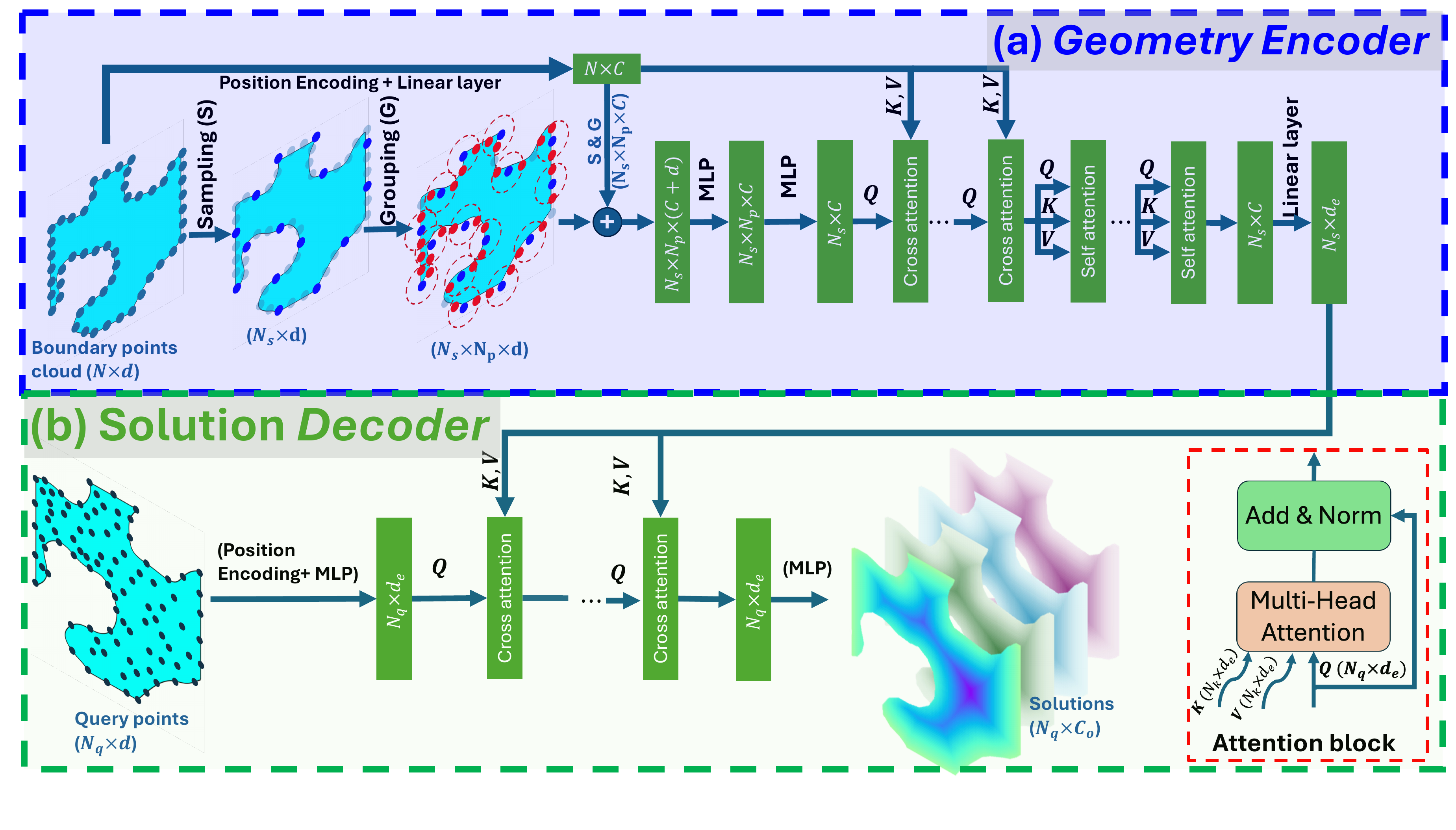}
    \caption{ Overview of the Geometry-Informed Neural Operator Transformer (GINOT)
    architecture. The boundary point cloud is initially processed through
    sampling and grouping layers to extract local geometric features. These local
    features are then fused with global geometric information via a cross-attention
    layer. This is followed by a series of self-attention layers and a final
    linear layer, producing the KEY and VALUE matrices for the cross-attention layer
    in the solution decoder. In the solution decoder, an multilayer perceptron (MLP)
    encodes the query points into the QUERY matrix for the cross-attention layer,
    which integrates the geometry information from the encoder. The output of the
    cross-attention layer is subsequently decoded into solution fields at the query
    points using another MLP. }
    \label{fig:ginot}
  \end{figure}

  GINOT integrates the transformer architecture with the neural operator
  framework to enable forward predictions for arbitrary geometries. It consists of
  two primary components: the geometry encoder and the solution decoder, as illustrated
  in \cref{fig:ginot}. Both components leverage the attention mechanism inherent
  in the transformer architecture \citep{Ashish2017atten}, which facilitates the
  model's ability to focus on relevant input information. The attention mechanism
  is defined as
  \begin{equation}
    \text{Attention}(Q,K,V) = \text{softmax}\left(\frac{QK^{T}}{\sqrt{d_{e}}}\right
    )V,
  \end{equation}
  where $Q \in \mathbb{R}^{N_q \times d_{e}}$ represents the QUERY matrix, and
  $K , V \in \mathbb{R}^{N_k \times d_{e}}$ are the KEY and VALUE matrices, respectively.
  Here, $d_{e}$ denotes the embedding dimension, $N_{q}$ is the length of the
  query sequence, and $N_{k}$ is the length of the KEY and VALUE sequences. To
  ensure stable training, residual connections and layer normalization are applied
  within each attention block, as depicted in the bottom right of
  \cref{fig:ginot}.

  In the following sections, we detail the GINOT architecture, omitting the
  batch size dimension for simplicity. For instance, we describe the points
  cloud as having dimensions $N \times d$, where $N$ represents the number of points
  and $d$ denotes the dimensionality. In practice, the actual shape of the points
  cloud is $B \times N \times d$, with $B$ being the batch size.

  \subsection{Geometry encoder}
  \label{sec:geo_encoder}

  Traditional convolutional architectures rely on highly regular input formats, such
  as image grids or 3D voxels, to efficiently perform kernel operations. While effective
  for structured data, these formats are unsuitable for capturing arbitrary
  geometries. In contrast, boundary points clouds offer a flexible
  representation for arbitrary geometries but lack the regularity required by
  convolutional architectures, presenting unique challenges for processing.

  To address these challenges, we developed a geometry encoder that processes
  the boundary point cloud of a geometry into the KEY and VALUE matrices for the
  attention mechanism. This enables seamless integration of geometry information
  with query points in the solution decoder. In contrast to CNN inputs, boundary
  point cloud used to represent geometries have three key characteristics: (1)
  invariance to the order of points, (2) non-uniform point density across locations,
  and (3) a variable number of points across geometries. Drawing inspiration from
  Shap-E \citep{jun2023shap} by OpenAI, the geometry encoder, as shown in
  \cref{fig:ginot}(a), is designed to handle these characteristics. It ensures invariance
  to point order and padding while maintaining robustness to variations in point
  density.

  The point cloud is first processed through a sampling layer and a grouping layer
  for local geometry features, originally introduced in PointNet \citep{qi2017pointnet}
  and PointNet++ \citep{qi2017pointnet++}, which are well-suited to handle the aforementioned
  properties. The sampling layer employs the iterative farthest point sampling (FPS)
  method \citep{qi2017pointnet++} to select $N_{s}$ points from the point cloud.
  This method begins by randomly selecting a point, then iteratively selects the
  farthest point from the already selected points until $N_{s}$ points are chosen.
  Compared to random sampling, the FPS method provides better coverage of the entire
  point set for a given number of centroids. The selected points are then passed
  to the grouping layer, which forms $N_{s}$ groups, each containing $N_{p}$
  points.

  For each group, the $N_{p}$ points are selected as those within a ball of radius
  $r$ (a hyperparameter) centered at the corresponding sampled point. If fewer
  than $N_{p}$ points exist within the ball, the group is padded with the nearest
  points to the center. Conversely, if more than $N_{p}$ points are present, the
  nearest $N_{p}$ points are chosen. Consequently, the output of the grouping
  layer has dimensions $N_{s}\times N_{p}\times d$, where $d$ represents the dimensionality
  of the point cloud. The sampling and grouping layers allow to extract local
  geometry features from nonuniform density point cloud \citep{qi2017pointnet++}.
  These layers output the selected points along with their indices in the
  original point cloud, which are subsequently used in downstream layers.

  The point cloud is also processed through a positional encoder layer utilizing
  the Nerf encoding method \citep{mildenhall2021nerf}, followed by a linear layer.
  Using the indices from the sampling and grouping layers, the output is indexed
  to extract local geometry information, resulting in a tensor of shape
  $N_{s}\times N_{p}\times C$, where $C$ represents the output channel dimension
  of the positional encoder. This tensor is concatenated with the output of the
  grouping layer. The combined tensor is then passed through a series multi-layer
  perceptron (MLP) layers, which aggregates the information within each group, reducing
  the dimension of $N_{p}$ to 1. As a result, the final output has a shape of
  $N_{s}\times C$.

  The local information obtained by the sampling and grouping layers is utilized
  as the QUERY ($Q$) in the cross-attention blocks, while the KEY ($K$) and VALUE
  ($V$) are derived from the output of the positional encoder layer, which
  captures global geometry information. The attention mechanism allows the local
  information in the QUERY ($Q$) to focus on the relevant global geometry features.
  The QUERY ($Q$) can be invariant to the permutation of points order based on
  the sampling and grouping mechanism, whereas the KEY ($K$) and VALUE ($V$) depend
  on the order of points as they are directly derived from the positional encoder
  layer with the point cloud as input. However, the output of the cross-attention
  block remains invariant to the order of points because of:
  \begin{equation}
    \text{Attention}(Q,K,V)_{id}= \text{softmax}\left(\frac{Q_{ij}K_{mj}}{\sqrt{d_{e}}}
    \right)V_{md},
  \end{equation}
  where the output is unaffected by permutations of the index $m$, which is a dummy
  index under Einstein summation notation. %
  {\color{\HighlightColor}%
  To ensure strict invariance, the sampling and grouping layers must also be invariant to the permutation of the point cloud. This requires consistently selecting the same initial point in the iterative FPS sampling layer and grouping points based on their distance to the centroids chosen by the sampling layer. However, as the sampling and grouping layers can sufficiently extract the geometry features regardless of the initial point selection, we use random initialization of the first point during FPS iteration in training. During inference, the first point is fixed as the first non-padding point of the point cloud to ensure consistent results. Our experiments demonstrate that this random initialization has a negligible impact on GINOT's output (as shown in \cref{sec:puc}).%

  The computational cost of the sampling and grouping layers is mainly attributed to the FPS and ball query operations. The FPS step, which iteratively selects the farthest point by computing distances between all pairs of points in the point cloud, has a complexity of $O(N^{2})$. The ball query step, which identifies points within a specified radius around each sampled point, involves a complexity of $O(N \times N_{s})$, where $N$ is the total number of points in the point cloud and $N_{s}$ is the number of sampled points. Importantly, both $N$ and $N_{s}$ are typically much smaller than the number of grid points and surface mesh points required in SDF-based methods, making the sampling and grouping layers in GINOT computationally more efficient, as further elaborated in \cref{sec:jeb}. %
  }%

  The sampling and grouping layers, combined with the attention mechanism, allow
  that the geometry encoder handle the point cloud with varying points density,
  while remaining robust to the permutation of points order. Since the number of
  points in the point cloud varies across different geometries, padding is
  typically required to standardize input sizes within a batch for deep learning
  models. To prevent padding points from influencing the geometry encoder's output,
  a masking mechanism is employed. This mechanism excludes padding points from being
  selected in the sampling and grouping layers. Additionally, padding points can
  interfere with attention scores in the cross-attention layer of the geometry encoder.
  To mitigate this, the corresponding entries in the attention score matrix $(Q \cdot
  K^{T})/\sqrt{d_{e}}$ are set to "$-\infty$", ensuring that the attention layer's
  output is unaffected by padding points. The flexibility and robustness of the proposed
  geometry encoder are illustrated in experiments \cref{sec:experiments}.

  \subsection{Solution decoder}
  \label{sec:sol_decoder}

  The solution decoder predicts the solution field at the query points using the
  geometry information from the geometry encoder. First, the query points are
  processed through a Nerf positional encoding, followed by an MLP, to generate
  the QUERY matrix with dimensions $N_{q}\times d_{e}$, where $N_{q}$ is the number
  of query points and $d_{e}$ is the embedding dimension. This QUERY matrix is
  then passed into cross-attention layers, which integrate the geometry information
  from the geometry encoder. The attention mechanism ensures that the query
  points focus on the most relevant geometry features. Finally, the output of
  the cross-attention layers is decoded into the solution field at the query
  points using another MLP. The architecture of the solution decoder is illustrated
  in \cref{fig:ginot}(b).

  Since the geometries are arbitrary, the number of query points varies across different
  geometries. To handle this variability, GINOT processes batches of geometries by
  padding the query points to match the maximum count within each batch, similar
  to techniques used for variable sequence lengths in natural language
  processing.

  \begin{figure}[htbp]
    \centering
    \includegraphics[width=6in]{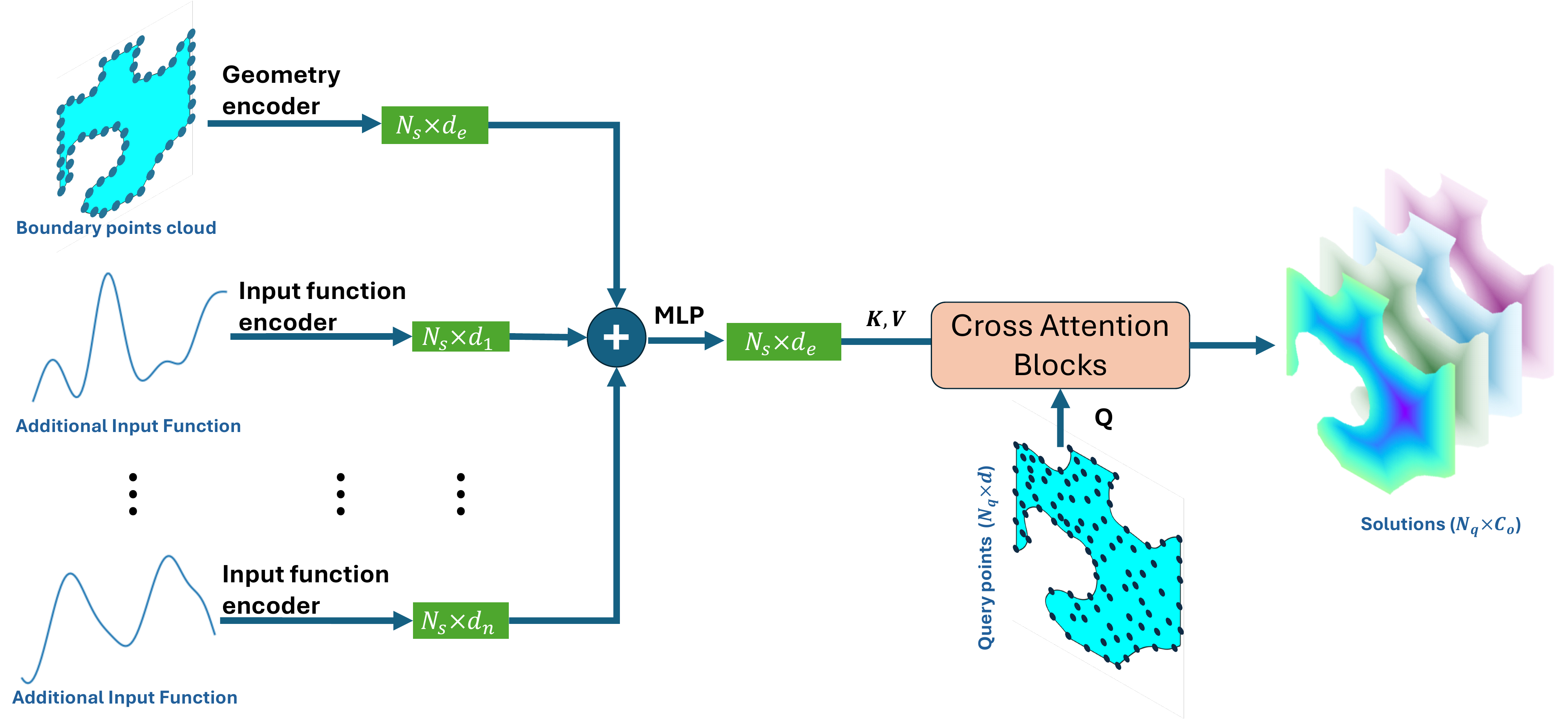}
    \caption{Extending GINOT for additional inputs. Encoders for extra inputs process
    these features, which are then concatenated with the geometry encoder's
    output. The combined representation is passed through an MLP for aggregation
    and used as the KEY and VALUE in the solution decoder's cross-attention blocks.
    This extension enables GINOT to incorporate both geometric and non-geometric
    inputs effectively.}
    \label{fig:x_ginot}
  \end{figure}

  \subsection{Extending GINOT}
  {\color{\HighlightColor}%
  Many problems involve not only varying geometries but also additional input parameters, such as loading conditions, material properties, or boundary conditions. To address such cases, GINOT can be extended by incorporating additional encoders to process these input functions. The outputs of these additional encoders are concatenated with the output of the geometry encoder and passed through an MLP for aggregation of all information from the geometries and the additional inputs. The aggregated output is then used as the KEY and VALUE in the cross-attention blocks of the solution decoder. The architecture of the extended GINOT is illustrated in \cref{fig:x_ginot}. This extension is demonstrated in the bracket lug example in \cref{sec:bracket_lug}, where varying loading conditions are included as additional inputs.%
  }%

  \section{Experiments}
  \label{sec:experiments}

  In this section, we present several numerical experiments to evaluate the performance
  and accuracy of GINOT on various challenging datasets. The primary evaluation
  metric used is the $L_{2}$ relative error defined as
  \begin{equation}
    L_{2}= \frac{\|y^{\text{true}}-y^{\text{pred}}\|_{2}}{\|y^{\text{true}}\|_{2}}
    , \label{eq:l2}
  \end{equation}
  where $y^{\text{true}}$ and $y^{\text{pred}}$ represent the ground truth and
  predicted solutions, respectively.

  To train the GINOT models, we use the mean square error (MSE) as the loss
  function. Since the number of query points is padded to match the maximum number
  of nodes in the batch, a masking mechanism is applied to exclude the padding
  points from the loss calculation:
  \begin{equation}
    \text{MSE}= \frac{1}{1 + \sum_{i=1}^{N}m_{i}}\sum_{i=1}^{N}m_{i}\left( y_{i}-
    \hat{y}_{i}\right)^{2},
  \end{equation}
  where $m_{i}$ is the mask value, set to 1 for non-padding points and 0 for
  padding points. The addition of 1 in the denominator ensures the denominator is
  never zero. Prior to training, both input and output data are normalized to have
  zero mean and unit variance.

  This section provides an overview of the datasets used in the experiments,
  along with the training hyperparameters and computational efficiency of GINOT
  for these datasets. Additionally, we summarize GINOT's performance on these
  datasets and compare it with baseline methods. Detailed experimental results are
  presented in the subsequent subsections. Further implementation details and
  results can be found in our
  \href{https://github.com/QibangLiu/GINOT}{GitHub repository}.

  \subsection{Overview}
  To evaluate the performance and accuracy of GINOT, six datasets are utilized:
  elasticity \citep{li2023fourier}, Poisson equation with both structured and
  unstructured meshes, bracket lugs \citep{he2024geom}, micro-periodic unit cell
  (PUC) \citep{liu2025dataset} in \citep{liu2025towards}, and Jet Engine Bracket
  (JEB) \citep{hong2024deepjeb}. \cref{tab:dataset} provides a detailed summary
  of these datasets, including their size, the number of query points, the number
  of points in the boundary point cloud, whether the dataset is parametric or
  fully arbitrary, and the type of solution considered. The first three datasets
  have a constant number of points in the boundary point cloud, while the last three
  datasets feature a varying number of points in the boundary point cloud. All
  datasets feature non-uniform point densities, as illustrated in the visualization
  results in the following subsections. %
  {\color{\HighlightColor} %
  While the datasets considered here focus on representative linear and nonlinear single-physics cases, the framework is not confined to single-physics or weakly coupled settings. With suitably generated training data, it works for complex, strongly coupled multi-physics problems. %
  } %

  \begin{table}[!h]
    \centering
    \caption{Dataset overview.}
    \begin{tabular}{c|cccccc}
      \hline
      DataSet                                                & Elasticity\citep{li2023fourier} & \makecell[cl]{Poisson \\ (structure mesh)} & \makecell{Poisson \\ (unstructure mesh)} & Bracket lug \citep{he2024geom} & \makecell{Micro-\\PUC \citep{liu2025dataset}} & JEB \citep{hong2024deepjeb} \\
      \hline
      Data size                                              & 2000                            & 6000                                       & 6000                                     & 3000                           & 73879                                         & 2138                        \\
      \cline{1-1} \makecell{Number of \\ query point}        & 972                             & 2737                                       & [1566,2838]                              & [5054,15210]                   & [1136,6363]                                   & [18085,55009]               \\
      \cline{1-1} \makecell{Number points \\ in boundary PC} & 105                             & 144                                        & 144                                      & [2184,4284]                    & [180,484]                                     & [2253,10388]                \\
      \cline{1-1} Parameteric                                & True                            & True                                       & True                                     & True                           & False                                         & False                       \\
      \cline{1-1} Solution                                   & Stress                          & --                                         & --                                       & Mises stress                   & \makecell[cl]{Mises stress \\ displacements}  & Mises stress                \\
      \hline
    \end{tabular}
    \label{tab:dataset}
  \end{table}

  \cref{tab:hyperpara} summarizes the training hyperparameters of the GINOT
  models. For the micro PUC dataset, two models are trained: one for predicting the
  solution at the final strain step and another for predicting the solution over
  historical strain steps.

  \begin{table}[!h]
    \centering
    \caption{Training hyperparameters of the GINOT models. All models are implemented
    with the PyTorch scheduler \emph{ReduceLROnPlateau}.}
    \begin{tabular}{c|ccccccc}
      \hline
                                  & Elasticity & \makecell{Poisson \\ (structure \\ mesh)} & \makecell{Poisson \\ (unstructure \\ mesh)} & \makecell{Bracket \\ lug} & \makecell{Micro PUC \\ (Final \\ strain step)} & \makecell{Micro PUC \\ (historical \\ strain steps)} & JEB    \\
      \hline
      Batch size                  & 20         & 32                                        & 32                                          & 32                        & 64                                             & 64                                                   & 16     \\
      Optimizer                   & Adam       & Adam                                      & Adam                                        & Adam                      & Adam                                           & Adam                                                 & AdamW  \\
      Initial learning rate       & 0.001      & 0.001                                     & 0.001                                       & 0.001                     & 0.001                                          & 0.001                                                & 0.0005 \\
      scheduler Patience          & 40         & 40                                        & 40                                          & 20                        & 10                                             & 10                                                   & 100    \\
      scheduler Factor            & 0.7        & 0.7                                       & 0.7                                         & 0.7                       & 0.7                                            & 0.7                                                  & 0.7    \\
      Epochs                      & 1000       & 500                                       & 500                                         & 500                       & 400                                            & 500                                                  & 500    \\
      Training dataset            & 1000       & 80\%                                      & 80\%                                        & 80\%                      & 80\%                                           & 80\%                                                 & 90\%   \\
      Testing dataset             & 200        & 20\%                                      & 20\%                                        & 20\%                      & 20\%                                           & 20\%                                                 & 10\%   \\
      $N_{s}$                     & 16         & 64                                        & 64                                          & 512                       & 128                                            & 128                                                  & 512    \\
      $N_{p}$                     & 18         & 18                                        & 18                                          & 64                        & 8                                              & 8                                                    & 64     \\
      Grouping $r$                & 0.2        & 0.2                                       & 0.2                                         & 0.5                       & 0.1                                            & 0.2                                                  & 0.1    \\
      Att. heads (Decoder)        & 4          & 8                                         & 8                                           & 8                         & 4                                              & 4                                                    & 4      \\
      Attention heads (Encoder)   & 4          & 8                                         & 8                                           & 8                         & 8                                              & 8                                                    & 1      \\
      Cross att. layers (Encoder) & 1          & 1                                         & 1                                           & 1                         & 2                                              & 2                                                    & 1      \\
      Self att. layers (Encoder)  & 2          & 3                                         & 3                                           & 2                         & 2                                              & 2                                                    & 2      \\
      Cross att. layers (Decoder) & 6          & 4                                         & 4                                           & 3                         & 4                                              & 4                                                    & 4      \\
      \hline
    \end{tabular}
    \label{tab:hyperpara}
  \end{table}

  \cref{tab:efficiency} summarizes the computational efficiency of GINOT. Training
  tasks were conducted on single NVIDIA A100 GPU on the DELTA machine or single NVIDIA
  H100 GPU on the DeltaAI machine, both hosted at the National Center for
  Supercomputing Applications (NCSA) at the University of Illinois Urbana-Champaign
  (UIUC). Inference tasks were executed on single NVIDIA A100 GPU on the DELTA machine.

  \begin{table}[!h]
    \centering
    \caption{GINOT computation efficiency.}
    \begin{tabular}{c|ccccccc}
      \hline
                             & Elasticity & \makecell{Poisson \\ (structure \\ mesh)} & \makecell{Poisson \\ (unstructure \\ mesh)} & \makecell{Bracket \\ lug} & \makecell{Micro PUC \\ (Final \\ strain step)} & \makecell{Micro PUC \\ (historical \\ strain steps)} & JEB    \\
      \hline
      Training GPU           & A100       & A100                                      & A100                                        & H100                      & H100                                           & H100                                                 & H100   \\
      Training [s/epoch]     & 1.7        & 12.1                                      & 11.9                                        & 33.2                      & 128                                            & 136                                                  & 31     \\
      Inference [s/solution] & 3.94e-4    & 1.03e-3                                   & 9.2e-4                                      & 1.35e-2                   & 2.18e-3                                        & 4.82e-3                                              & 4.0e-2 \\
      \hline
    \end{tabular}
    \label{tab:efficiency}
  \end{table}

  \cref{tab:perfom} summarizes GINOT's performance across the datasets, presenting
  the $L_{2}$ relative error for both training and testing samples. These
  results are compared against baseline methods, with the best performance highlighted
  in bold. The findings demonstrate GINOT's ability to tackle more complex
  problems while outperforming the baseline methods. Detailed experimental results
  are provided in the subsequent subsections.

  \begin{table}[!h]
    \centering
    \caption{Performance comparison of GINOT with baseline methods. ``--" indicates
    that the method cannot handle the dataset, while ``NA" signifies that the method
    can handle the dataset but results are unavailable. The best performance is
    highlighted in bold.}
    \begin{tabular}[c]{l|lccccc}
      \hline
      Dataset, Metric                                                                       & Data     & Geo-FNO\citep{li2023fourier} & \makecell{Geom-\\DeepONet\citep{he2024geom}} & GNO\citep{li2020neural} & GI-FNO\citep{li2023GIFNO} & GINOT             \\
      \hline
      \multirow{2}{*}{Elasticity, $L_{2}$}                                                  & Training & 1.25\%                       & NA                                           & 11.3\%                  & 1.32\%                    & \textbf{0.806\%}  \\
                                                                                            & Testing  & 2.29\%                       & NA                                           & 12.3\%                  & 1.66\%                    & \textbf{1.03\%}   \\
      \cline{1-1} \multirow{2}{*}{\makecell{Poisson, $L_{2}$ \\ (structure mesh)} }         & Training & 0.43\%                       & NA                                           & 7.43\%                  & 0.42\%                    & \textbf{0.39\%}   \\
                                                                                            & Testing  & 0.48\%                       & NA                                           & 7.82\%                  & 0.45\%                    & \textbf{0.437\%}  \\
      \cline{1-1} \multirow{2}{*}{\makecell{Poisson, $L_{2}$ \\ (unstructure mesh)} }       & Training & --                           & NA                                           & 7.98\%                  & 0.49\%                    & \textbf{0.41\%}   \\
                                                                                            & Testing  & --                           & NA                                           & 8.23\%                  & 0.55\%                    & \textbf{0.53\%}   \\
      \cline{1-1} \multirow{2}{*}{Bracket lug,$L_{2}$}                                      & Training & --                           & 1.818\%                                      & 10.5\%                  & 2.51\%                    & \textbf{0.35\%}   \\
                                                                                            & Testing  & --                           & 1.824\%                                      & 11.3\%                  & 2.81\%                    & \textbf{0.40\%}   \\
      \cline{1-1} \multirow{2}{*}{\makecell{Micro PUC, $L_{2}$ \\(Final strain step)}}      & Training & --                           & --                                           & 16.5\%                  & 9.24\%                    & \textbf{6.99\%}   \\
                                                                                            & Testing  & --                           & --                                           & 18.2\%                  & 9.95\%                    & \textbf{7.61\%}   \\
      \cline{1-1} \multirow{2}{*}{\makecell{Micro PUC, $L_{2}$\\(historical strain steps)}} & Training & --                           & --                                           & 18.9\%                  & 9.56\%                    & \textbf{8.42\%}   \\
                                                                                            & Testing  & --                           & --                                           & 19.5\%                  & 10.01\%                   & \textbf{9.06\%}   \\
      \cline{1-1} \multirow{2}{*}{JEB, $L_{2}$}                                             & Training & --                           & --                                           & 36.4\%                  & 27.5\%                    & \textbf{18.77\% } \\
                                                                                            & Testing  & --                           & --                                           & 67.3\%                  & 31.4\%                    & \textbf{30.77\% } \\
      \hline
    \end{tabular}
    \label{tab:perfom}
  \end{table}

  \subsection{Elasticity}

  \begin{figure}[h]
    \centering
    \includegraphics[width=\textwidth]{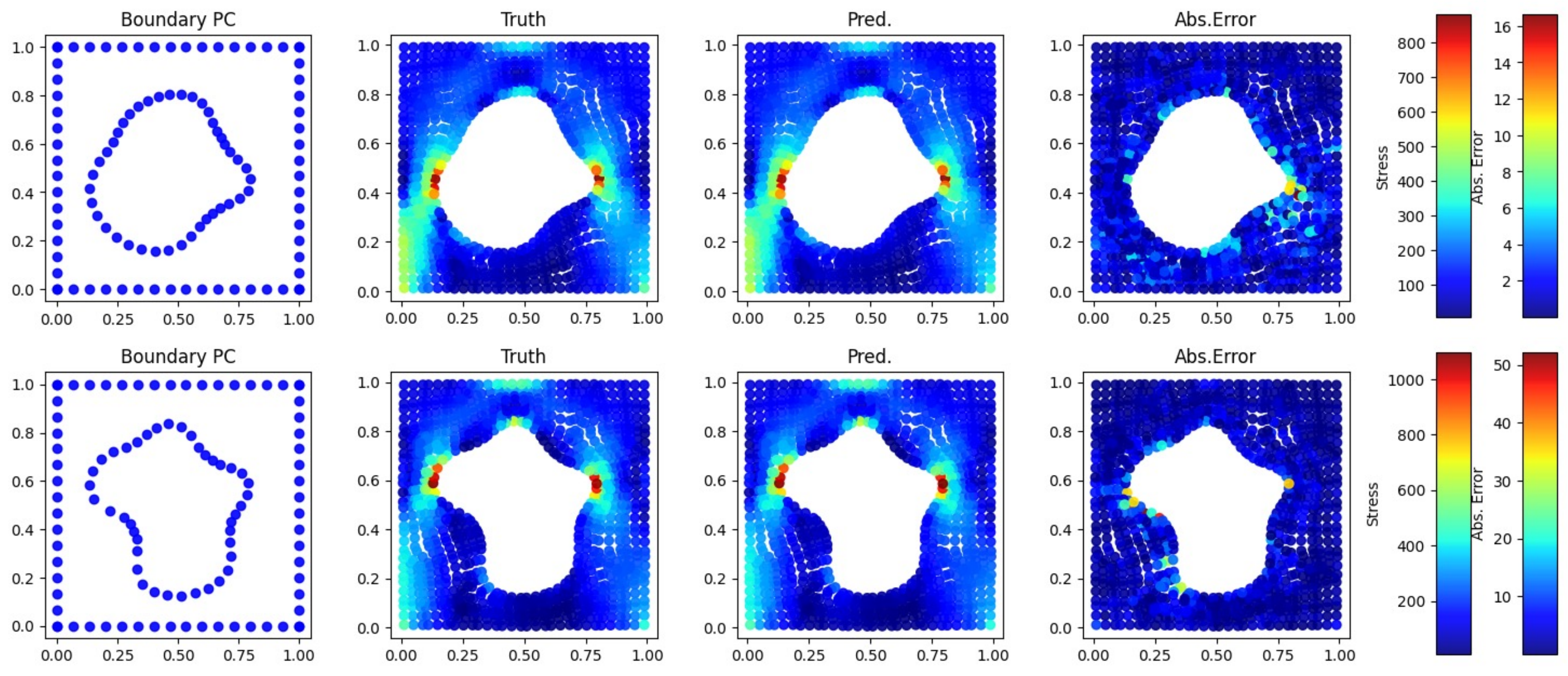}
    \caption{Stress solutions of the testing sample for the elasticity problem. The
    first row shows the median case ($L_{2}=1.23\%$), while the second row
    represents the worst case ($L_{2}=2.93\%$) based on $L_{2}$ relative error. The
    first column depicts the input surface point cloud, the second column
    presents the ground truth from finite element analysis, the third column shows
    the prediction by GINOT, and the last column highlights the absolute error
    between the prediction and the ground truth.}
    \label{fig:elasticity}
  \end{figure}

  For the elasticity example, we utilized the dataset generated by \citet{li2023fourier}
  for training the Geo-FNO model. The dataset consists of unit cells of $[0,1]^{2}$
  with arbitrary-shaped voids at the center, as illustrated in
  \cref{fig:elasticity}. The central void is defined by 45 points evenly distributed
  along the tangential direction, parameterized as
  \begin{equation}
    \begin{aligned}
      x_{i} & = r_{i}\cos(\theta_{i}), \\
      y_{i} & = r_{i}\sin(\theta_{i}),
    \end{aligned}
    \label{eq:void}
  \end{equation}
  where $r_{i}$ is the radius of the $i$-th point, and $\theta_{i}= (2\pi i)/45$.
  The radii $r_{i}$ follow a Gaussian distribution, constrained within $0.2 \leq
  r_{i}\leq 0.4$. The unit cell is fixed at the bottom and a tensile traction is
  applied at the top to simulate the stress field. The dataset uses a structured
  mesh with a fixed number of 972 nodes across all samples. A total of 1000
  training samples were used to train the model for predicting stress solutions
  for various geometries.

  Geo-FNO \citep{li2023fourier} employs 45 parametric void radii as geometry features,
  which limits its applicability to parametric geometries. In contrast, GINOT
  uses boundary points clouds to represent geometry, enabling it to handle arbitrary
  geometries. The stress solutions of the testing samples predicted by GINOT are
  shown in \cref{fig:elasticity}, highlighting the median and worst cases in terms
  of $L_{2}$ relative error. For the median case, the $L_{2}$ relative error is 0.96\%,
  while the worst case has an $L_{2}$ relative error of 2.31\%. The mean $L_{2}$
  relative error across the testing samples is 1.03\%, with a standard deviation
  of 0.35\%.

  \subsection{Poisson equation}

  \begin{figure}[h]
    \centering
    \includegraphics[width=\textwidth]{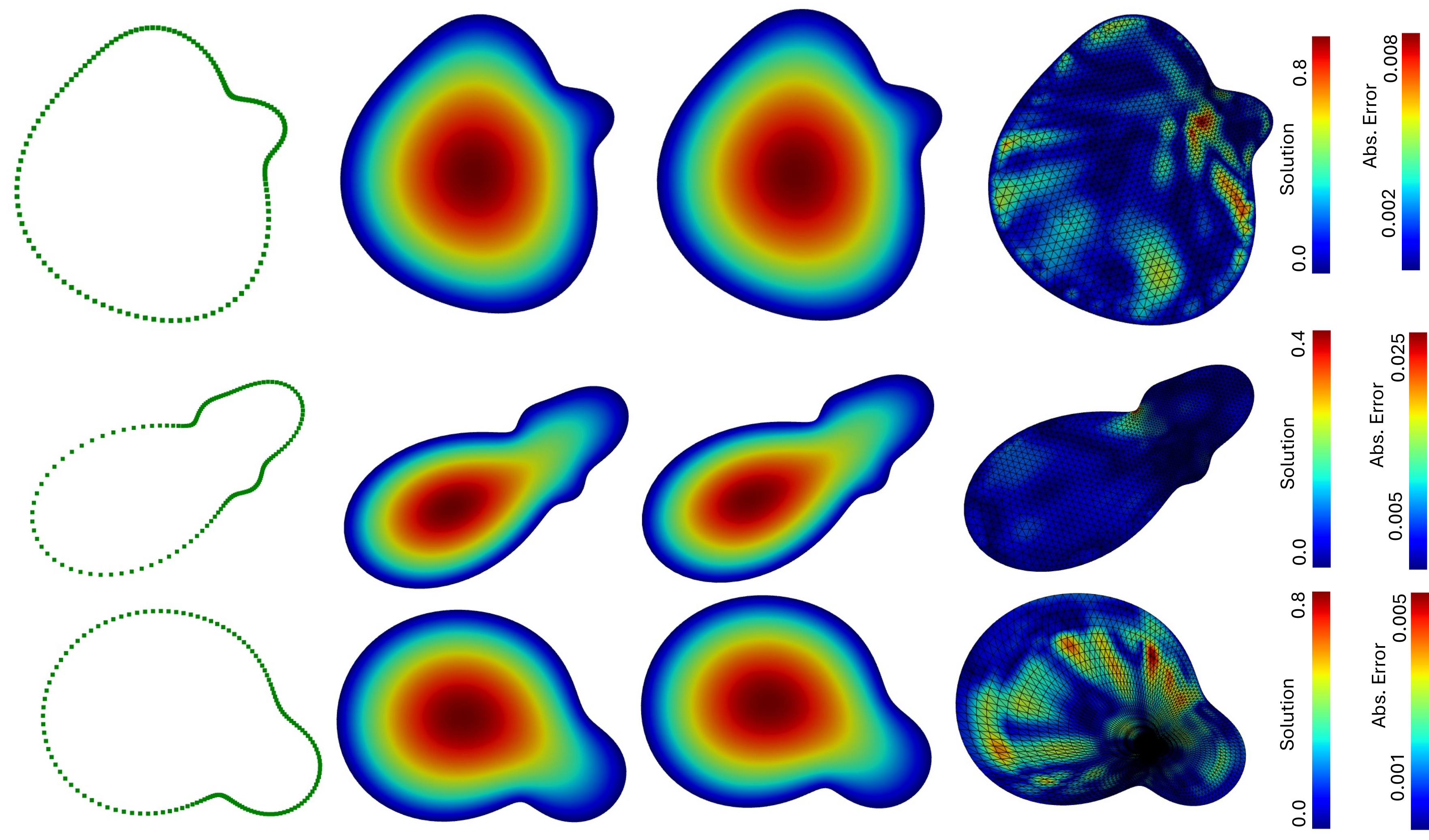}
    \caption{Testing sample solutions for the Poisson equation. The first row shows
    the median case ($L_{2}=0.41\%$) while the second row presents the worst case
    ($L_{2}=2.27\%$) for unstructured mesh data, based on $L_{2}$ relative error.
    The third row displays the median case ($L_{2}=0.38\%$) for structured mesh data.
    The first column illustrates the input surface point cloud, the second column
    shows the FEM ground truth, the third column provides GINOT predictions, and
    the last column highlights the absolute error between predictions and the ground
    truth. Structured and unstructured meshes are also depicted in the last
    column.}
    \label{fig:poisson}
  \end{figure}

  The Poisson equation considered in this study is given by
  \begin{equation}
    \begin{aligned}
      -\nabla^{2}u & = f, \quad \text{in}\quad \Omega,          \\
      u            & = g, \quad \text{on}\quad \partial \Omega,
    \end{aligned}
    \label{eq:poisson}
  \end{equation}
  where $\Omega$ represents the domain, $u$ is the solution, $f$ is the source
  term, and $g$ is the Dirichlet boundary condition. For all samples, we set $f=1$
  and $g=0$. Similar to the elasticity example, the domain $\Omega$ is parameterized
  using \cref{eq:void}, with the radius $r_{i}\in[0.2, 0.8]$ and
  $i=1,2,\ldots,14 4$. The geometries are illustrated in \cref{fig:poisson}.

  Two datasets were generated for this problem, each containing 6000 samples. The
  first dataset uses a structured mesh with a fixed 2737 nodes across all samples,
  while the second dataset employs an unstructured mesh, where the number of nodes
  varies between 1566 and 2838. The structured and unstructured meshes are illustrated
  in the last column of \cref{fig:poisson}. For comparison with Geo-FNO, which does
  not support unstructured meshes with varying node counts, we focus on the
  structured mesh dataset. The Poisson equation was solved using FEM with FEniCSX.
  Of the total samples, 20\% were reserved for testing, and the remaining 80\%
  were used for training.

  The comparison between GINOT predictions and the ground truth is presented in
  \cref{fig:poisson}. For the unstructured mesh dataset, the median case has an
  $L_{2}$ relative error of 0.41\%, while the worst case has an $L_{2}$ relative
  error of 2.27\%. For the structured mesh dataset, only the median case is shown
  for brevity, with an $L_{2}$ relative error of 0.38\%. The mean $L_{2}$
  relative error for the testing samples of the structured mesh dataset is 0.45\%,
  with a standard deviation of 0.17\%, while for the unstructured mesh dataset, the
  mean $L_{2}$ relative error is 0.44\%, with a standard deviation of 0.21\%.

  \subsection{Bracket lug}
  \label{sec:bracket_lug}

\begin{figure}[h]
    \centering
    \includegraphics[width=\textwidth]{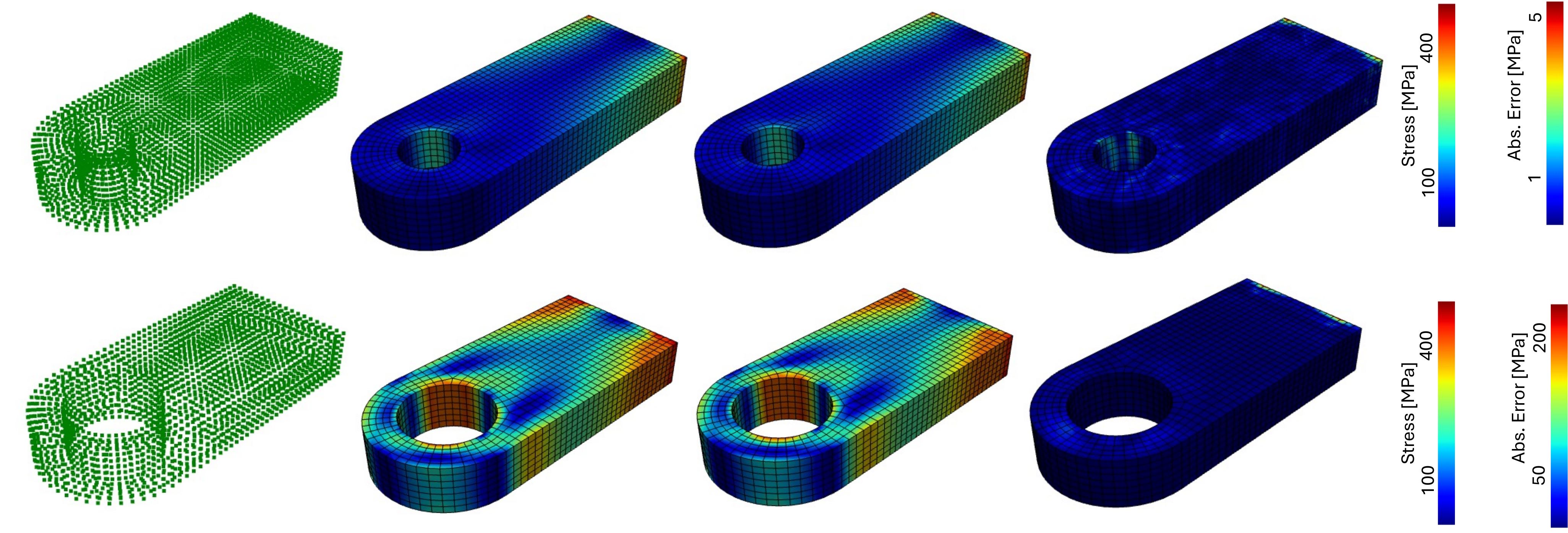}
    \caption{Stress solutions for the testing samples of bracket lugs. The first
    row shows the median case ($L_{2}=0.41\%$), while the second row represents
    the worst case ($L_{2}=3.78\%$) based on $L_{2}$ relative error. The first column
    illustrates the input surface point cloud, the second column presents the
    ground truth from finite element analysis, the third column shows the GINOT prediction,
    and the last column highlights the absolute error between the prediction and
    the ground truth.}
    \label{fig:lugs}
  \end{figure}

  We address the plasticity problem of bracket lugs using the dataset used in \citep{he2024geom}.
  The bracket lugs, depicted in \cref{fig:lugs}, are parameterized by three variables:
  length, thickness, and the radius of the circular hole. FEM simulations employ
  an elastic-plastic material model with linear isotropic hardening and quadratic
  hexahedral elements under small deformation assumptions. For training GINOT, the
  middle edge nodes of the quadratic hexahedral elements are removed. The
  surface mesh is extracted from the volume hexahedral elements, and its nodes are
  used as the boundary point cloud, which is then input to the geometry encoder
  of GINOT.

  { \color{\HighlightColor}%
  The dataset also includes variations in pressure loading across samples. To incorporate the loading parameter, we adopt the framework shown in \cref{fig:x_ginot}. An MLP encodes the loading (shape of $B \times 1$) into a tensor with shape of $B \times d_{e}$, which is expanded to $B \times N_{s}\times d_{e}$ by repeating the tensor $N_{p}$ times, where $B$ is the batch size. This tensor is concatenated with the geometry encoder's output and passed through an MLP, whose output serves as the KEY and VALUE for the attention mechanism. The dataset contains 3000 samples, with 20\% reserved for testing.%
  }

  The stress solutions for the testing samples predicted by GINOT are shown in \cref{fig:lugs},
  highlighting the median and worst cases based on $L_{2}$ relative error. The median
  case has an $L_{2}$ relative error of 0.41\%, while the worst case reaches 3.78\%.
  The mean $L_{2}$ relative error for the testing samples is 0.45\%, with a standard
  deviation of 0.23\%.

  \subsection{Micro-periodic unit cell}
  \label{sec:puc}

  \begin{figure}[h]
    \centering
    \includegraphics[width=\textwidth]{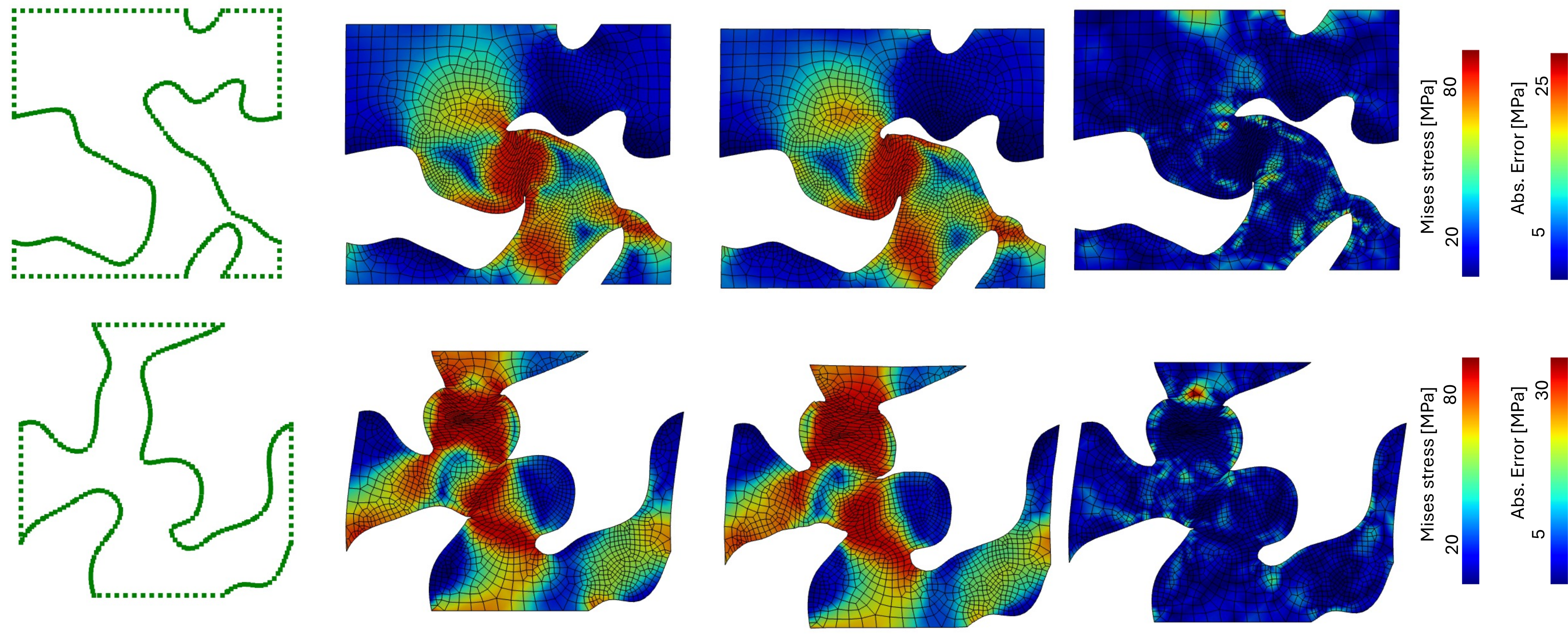}
    \caption{Mises stress and displacement solutions at the final strain step
    for the testing sample of the micro-periodic unit cell. The first row shows the
    median case ($L_{2}=7.54\%$), while the second row represents the 95th
    percentile case ($L_{2}=13.03\%$) based on $L_{2}$ relative error. The first
    column illustrates the input surface point cloud, the second column displays
    the true stress on the true deformed shape, the third column shows the predicted
    stress on the predicted deformed shape, and the last column highlights the
    absolute error of stress on the true deformed shape.}
    \label{fig:puc_laststep_plot}
  \end{figure}

  \begin{figure}[h]
    \centering
    \includegraphics[width=\textwidth]{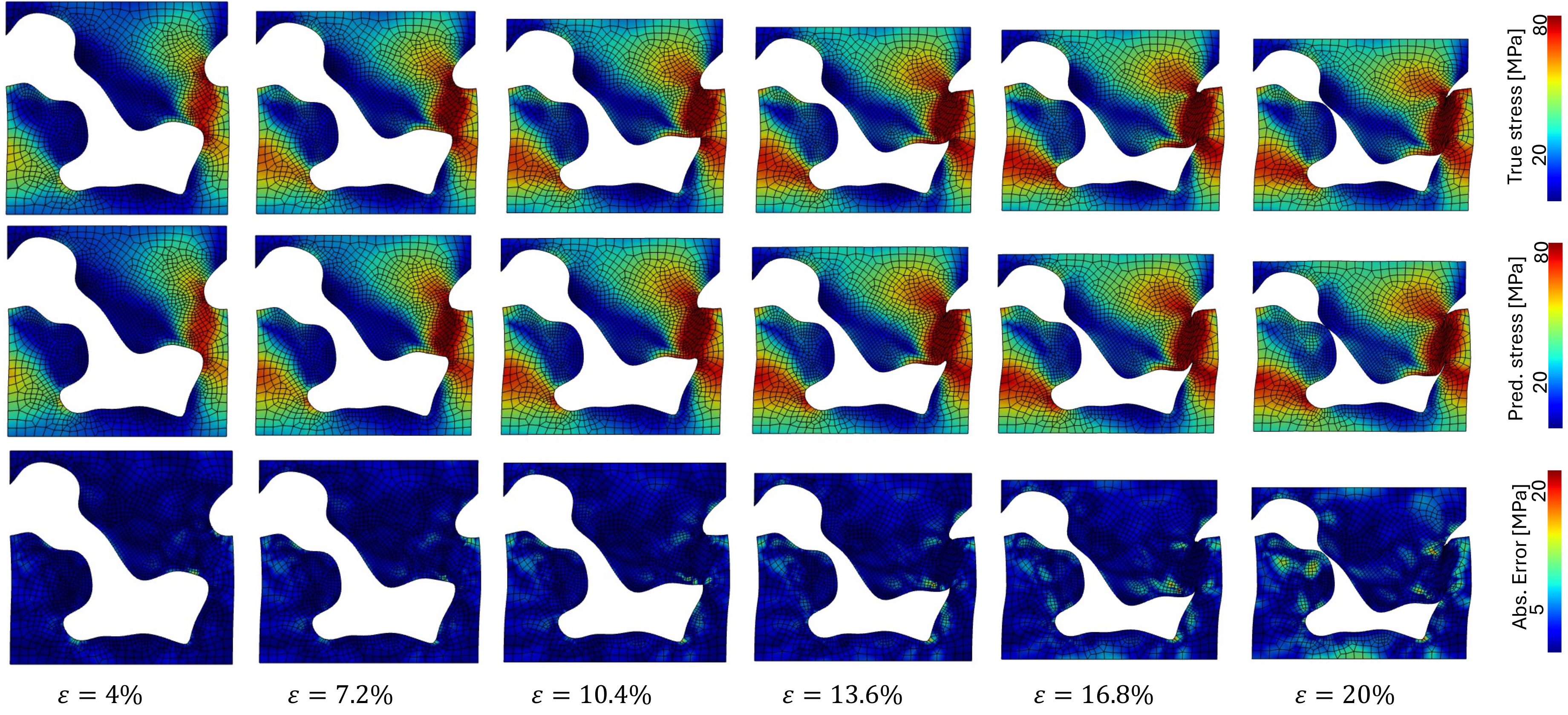}
    \caption{Mises stress and displacement solutions across historical strain
    steps for the median testing sample ($L_{2}=9.08\%$) of the micro-periodic
    unit cell, based on $L_{2}$ relative error. The first row shows the true
    stress on the true deformed shape, the second row presents the predicted stress
    on the predicted deformed shape, and the last row highlights the absolute
    error of the stress on the true deformed shape.}
    \label{fig:puc_26steps}
  \end{figure}

  \begin{figure}[h]
    \centering
    \includegraphics[width=\textwidth]{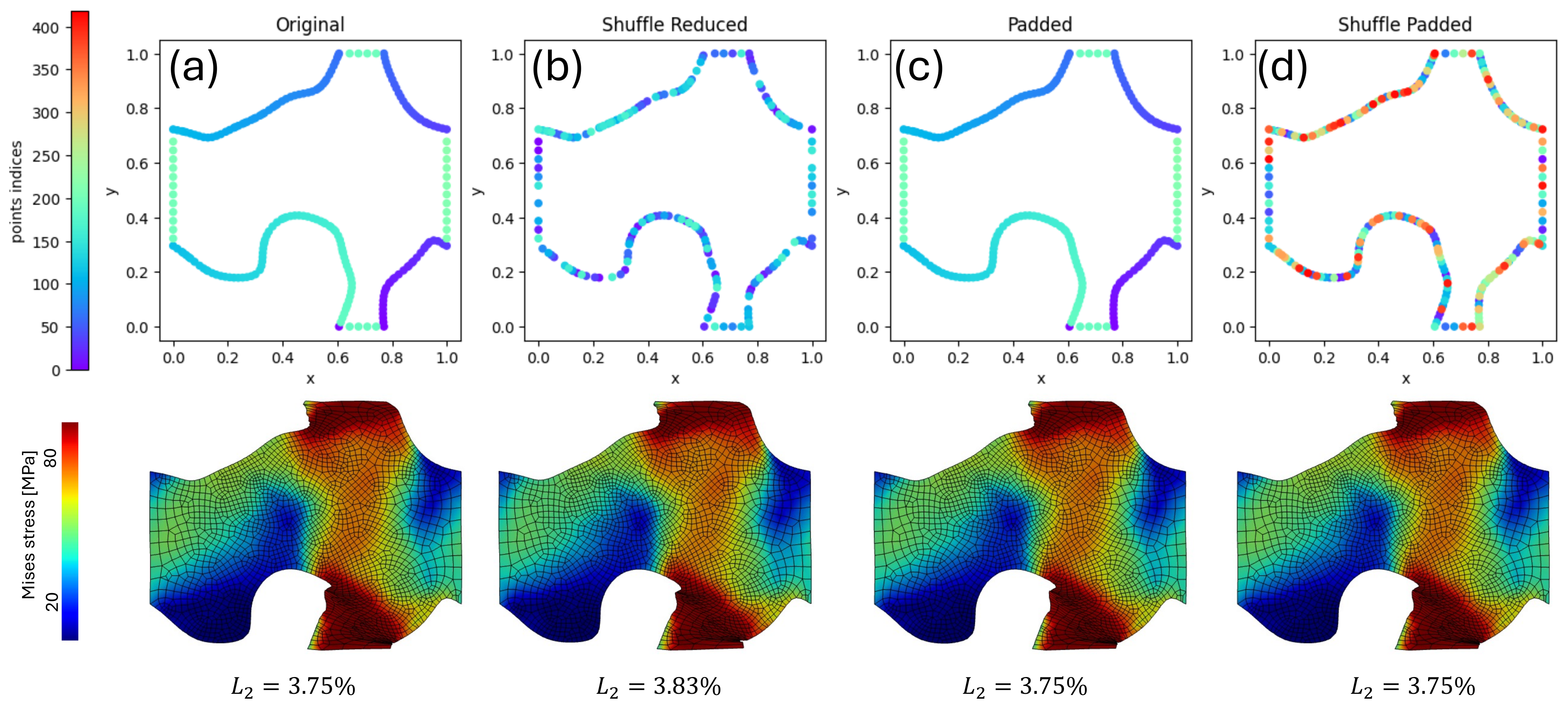}
    \caption{Demonstrating the versatility of the geometry encoder. The first row
    illustrates the input surface point cloud, with colors indicating the point
    order. The second row presents the corresponding predicted Mises stress on the
    predicted deformed shape by GINOT. (a) Original surface point cloud (PC)
    with 220 points, (b) shuffled original PC with the first 80\% retained, (c) original
    PC padded with 199 points at coordinates (-1000, -1000) (not shown), and (d)
    shuffled version of the PC in (c). The $L_{2}$ relative errors for each case
    are displayed at the bottom.}
    \label{fig:versa_geo_encoder}
  \end{figure}

  The previous examples focus on 2D or 3D geometries with relatively simple shapes
  parameterized by a few variables. In this example, we consider micro-periodic unit
  cells (PUC) generated by the authors for metamaterial architecture design.
  These geometries are more complex and fully arbitrary. The dataset is created by
  first generating a 2D periodic Gaussian random field on a square domain $[0,1~\text{mm}
  ]^{2}$. The Marching Squares algorithm is then applied to extract the contour
  at a random threshold value, defining the boundary of the 2D periodic unit cell,
  with field values below the threshold treated as void.

  Periodic boundary conditions and compression are applied to the unit cell for
  FEM simulations to obtain the mechanical response field. The simulations use an
  elasto-plastic material model with large deformation within the environment of
  Abaqus. To approximate quasi-static compression while improving convergence
  for large deformation and elasto-plastic behavior, an implicit dynamic solver is
  employed with a virtual mass density of $\rho=10^{-8}$. The compression strain
  is applied smoothly over 51 steps from $\varepsilon=0$ to $\varepsilon=20\%$. The
  simulation utilizes a combination of mixed first-order quadrilateral elements and
  second-order triangular elements with full integration. To augment the dataset,
  the unit cell is randomly shifted multiple times. The corresponding stress and
  displacement solution fields are also shifted accordingly, enabling data augmentation
  without additional FEM simulations. The boundary point cloud of the geometry
  is formed by combining the points on the contours from the Marching Squares
  algorithm and the points on the boundary of the square domain, spaced at intervals
  of $\frac{1}{32}$ mm. This boundary point cloud represents the geometry
  features and is fed into the geometry encoder.

  The dataset comprises around 73,879 samples, with 20\% reserved for testing. The
  GINOT model is trained to predict Mises stress and two displacement components,
  both at the final strain step and over historical strain steps.

  The predicted solutions at the final strain step are shown in
  \cref{fig:puc_laststep_plot}, comparing the ground truth with GINOT predictions.
  The first row highlights the median case with an $L_{2}$ relative error of
  7.54\%, while the second row shows the 95th percentile case with an $L_{2}$
  relative error of 13.03\%. The first column illustrates the input surface
  point cloud, the second column presents the true stress on the true deformed shape,
  the third column displays the predicted stress on the predicted deformed shape,
  and the last column shows the absolute error of stress on the true deformed shape.
  The mean $L_{2}$ relative error across the testing samples is 9.05\%, with a
  standard deviation of 3.53\%.

  The stress and displacement solutions over historical strain steps are compared
  with the ground truth and illustrated in \cref{fig:puc_26steps}. For brevity, only
  the median case is presented here. Additional cases, such as the best and worst,
  can be found in our
  \href{https://github.com/QibangLiu/GINOT}{Github repository}. The first row shows
  the true stress on the true deformed shape, the second row depicts the
  predicted stress on the predicted deformed shape, and the last row highlights the
  absolute error of stress on the true deformed shape. The overall $L_{2}$
  relative error for historical steps is 9.08\%. The mean $L_{2}$ relative error
  for the testing samples is 9.06\%, with a standard deviation of 3.02\%.

  As detailed in \cref{sec:geo_encoder}, the geometry encoder in GINOT is
  designed to handle 1) nonuniform point cloud densities, 2) permutations in point
  order, and 3) padding within the point cloud. The first property is evident
  throughout the examples in this work. To demonstrate the latter two properties,
  we analyze the second-best test case in this example. Notably, GINOT maintains
  stable inference performance even when the point cloud is made sparser. In this
  analysis, the point cloud is shuffled, reduced to 80\% of its original size, padded
  with 199 points at $(-1000, -1000)$, and shuffled again. These modified point clouds
  are input to the trained GINOT, with results shown in
  \cref{fig:versa_geo_encoder}. The $L_{2}$ relative errors for the original,
  reduced and shuffled, padded, and shuffled padded point clouds are 3.75\%,
  3.83\%, 3.75\%, and 3.75\%, respectively. The similar $L_{2}$ errors across these
  cases demonstrate the geometry encoder's robustness to point order and padding.
  {\color{\HighlightColor}%
  We emphasize that the results in \cref{fig:versa_geo_encoder} were produced using sampling and grouping layers implemented to be invariant to point permutations as described in \cref{sec:geo_encoder}; consequently, the padded and shuffled-padded variants yield identical $L_{2}$ errors (3.75\%). If a simpler FPS implementation is used that initializes from the first non-padding point (default of GINOT), the shuffled-padded case shows an increase in $L_{2}$ of only 0.002 relative to the padded case, indicating negligible impact.%
  } %
  Although shuffling and padding have minimal impact on model performance, a
  significant reduction in point cloud density can lead to an increase in the
  $L_{2}$ relative error due to the loss of geometric information. The effect of
  point cloud density on model performance is summarized in \cref{tab:point_density}.

  \begin{table}[h]
    \centering
    \caption{Impact of point cloud density on the inference performance of the
    trained model.}
    \begin{tabular}{c|ccccc}
      \hline
      \makecell{Point cloud density \\ (percentage of original)} & 100\%  & 80\%   & 60\%   & 40\%   & 20\%    \\
      \hline
      $L_{2}$ relative error                                     & 3.75\% & 3.83\% & 4.50\% & 8.18\% & 30.46\% \\
      \hline
    \end{tabular}
    \label{tab:point_density}
  \end{table}

  \begin{figure}[h]
    \centering
    \includegraphics[width=\textwidth]{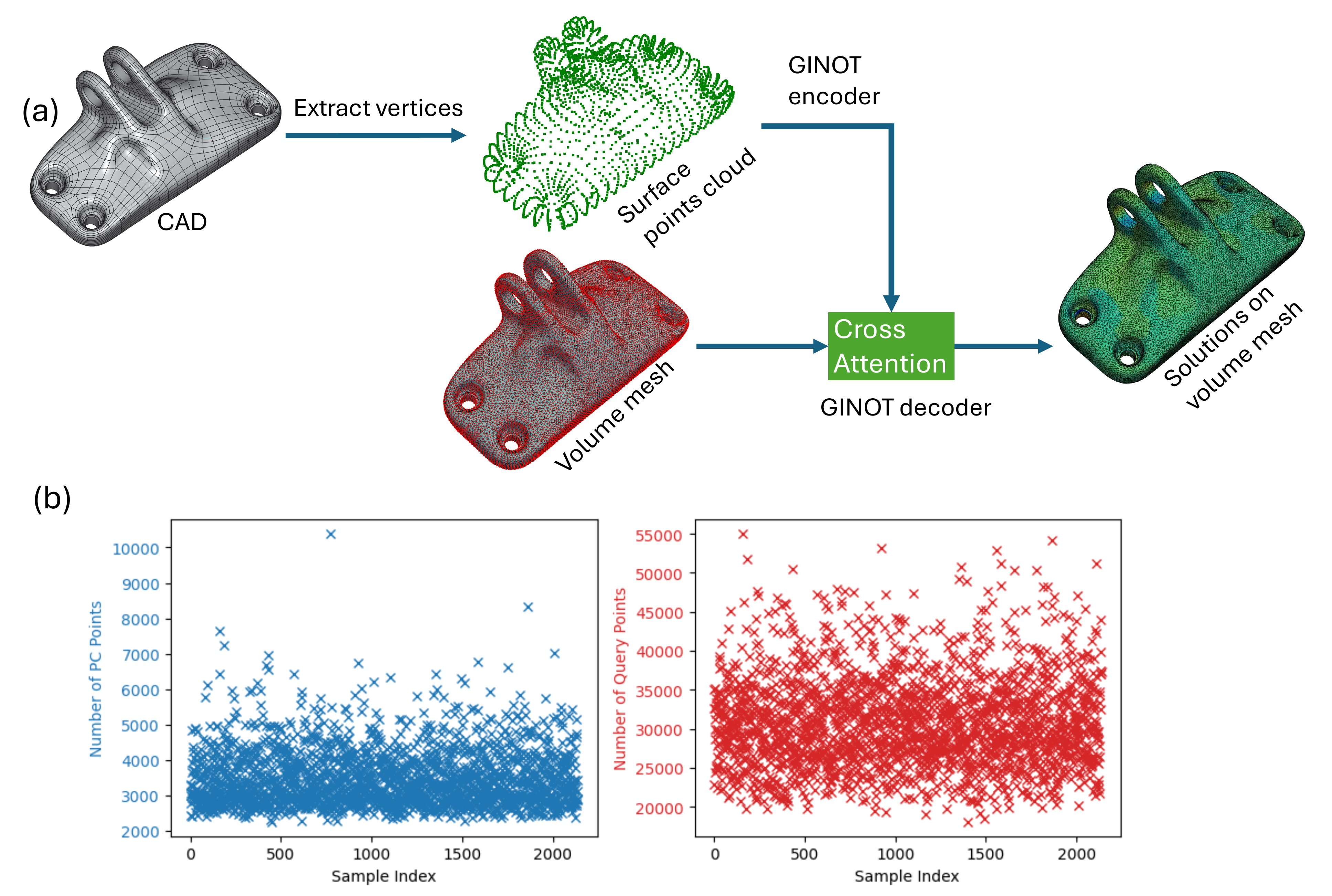}
    \caption{Workflow for the Jet Engine Bracket (JEB) dataset. (a) Boundary points
    clouds are extracted from the CAD models of each JEB sample using FreeCAD with
    PythonAPI and serve as input to the geometry encoder of GINOT. The 3D volume
    mesh nodes are utilized as query points and provided to the solution decoder
    of GINOT. (b) Distribution of the number of boundary points in the cloud and
    volume mesh nodes across the 2138 JEB samples.}
    \label{fig:jeb_data}
  \end{figure}

  \begin{figure}[!h]
    \centering
    \includegraphics[width=3in]{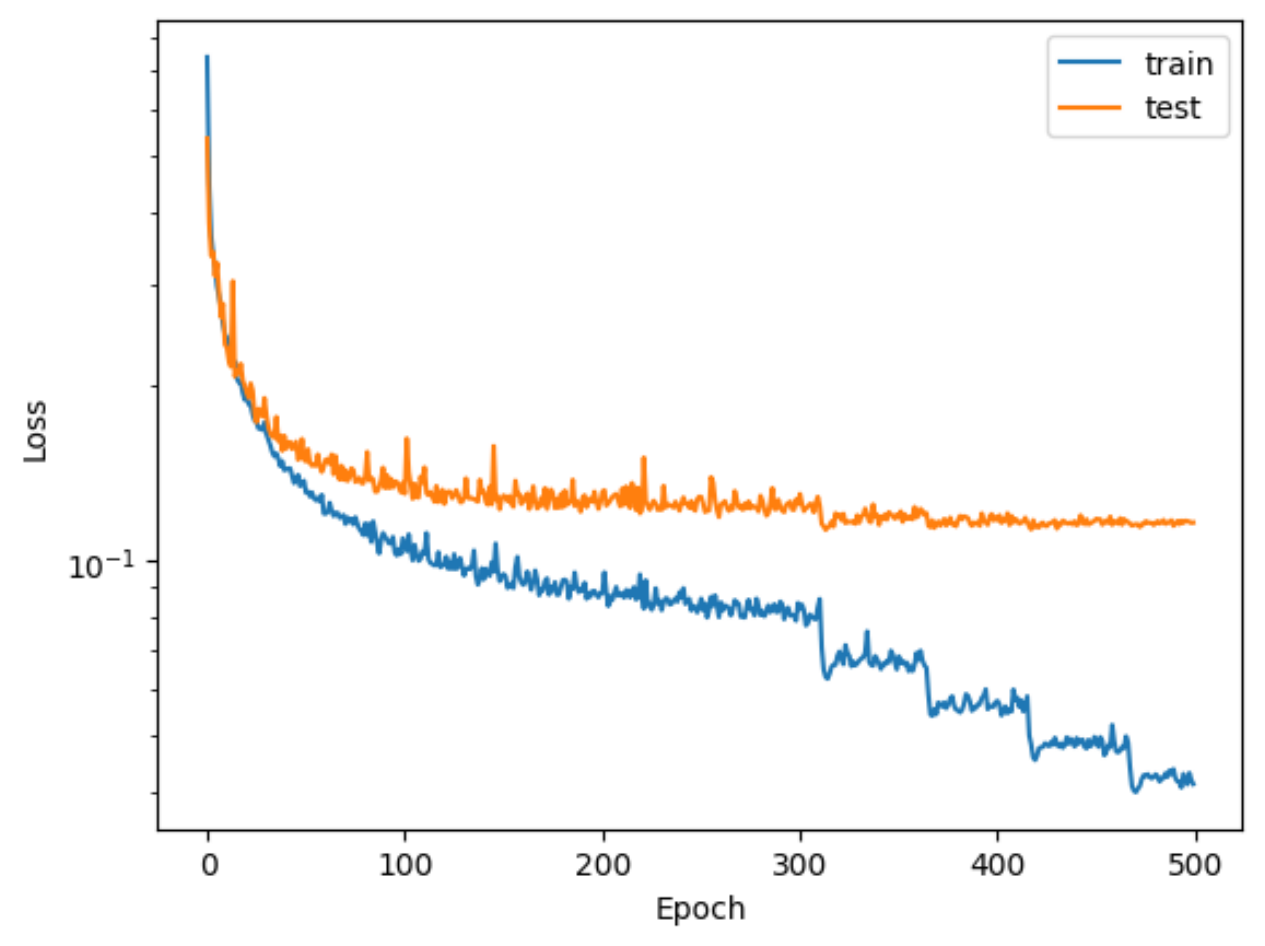}
    \caption{Training and validation mean squared error (MSE) loss curves for
    GINOT on the JEB dataset.}
    \label{fig:jeb_loss}
  \end{figure}

  \subsection{Jet engine bracket}
  \label{sec:jeb}

  The final example demonstrating GINOT's capability is the \href{https://www.narnia.ai/dataset}{Jet
  Engine Bracket (JEB)} dataset \citep{hong2024deepjeb}. This dataset comprises 2,138
  samples of jet engine brackets derived from the "GE Jet Engine Bracket
  Challenge" guidelines. These geometries are complex 3D structures with
  arbitrary shapes and sizes. FEM simulations provide solution fields (e.g.,
  stress, displacement) under four loading scenarios: vertical, horizontal, diagonal,
  and torsional, all with fixed loading magnitudes. In this example, we focus on
  the Mises stress solution under vertical loading.

  \begin{figure}[h]
    \centering
    \includegraphics[width=\textwidth]{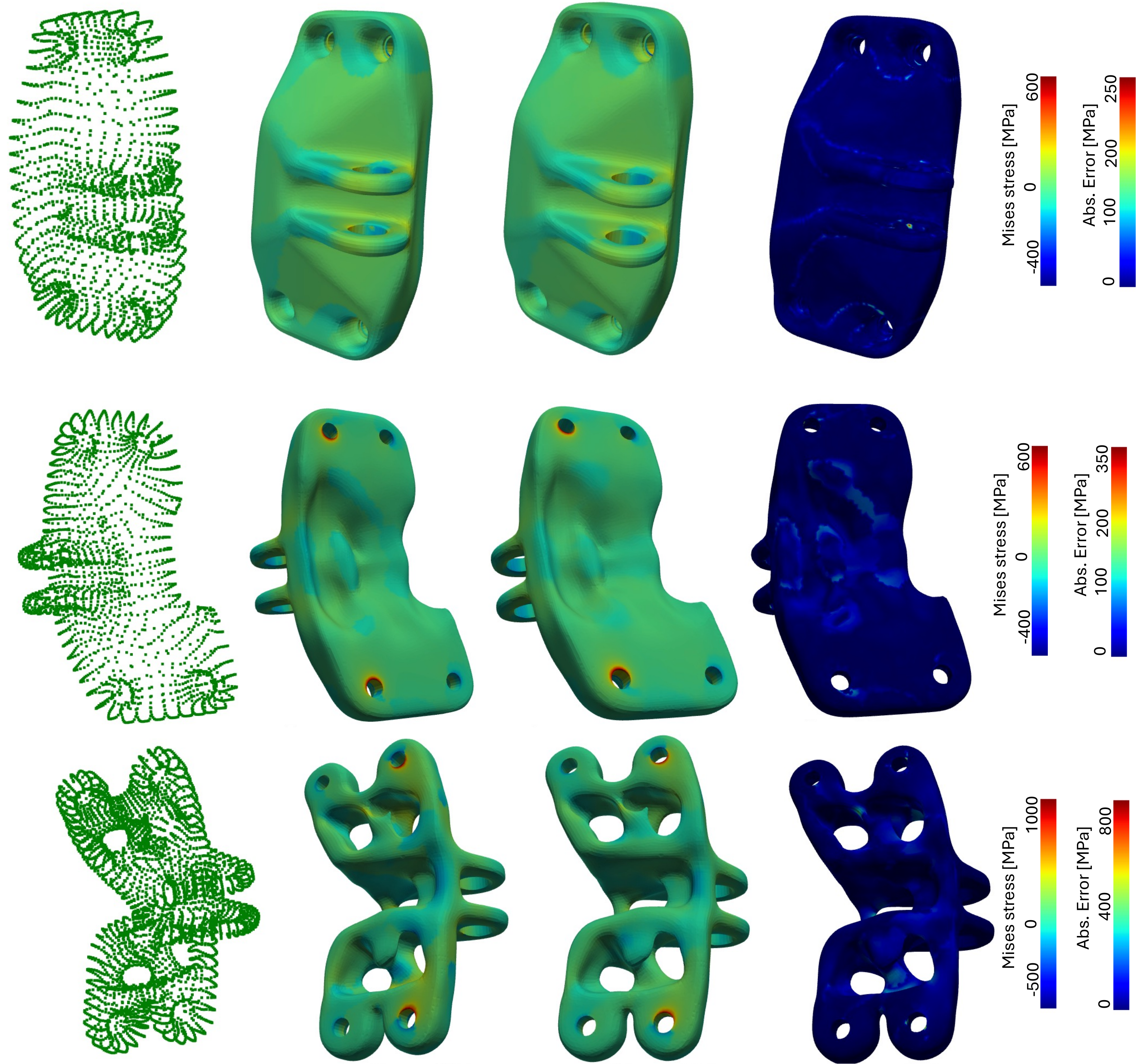}
    \caption{Mises stress solutions for the testing samples of the JEB dataset. The
    first row shows the best case ($L_{2}= 16.99\%$), the second row represents
    the median case ($L_{2}=29.0\%$), and the third row depicts the worst case ($L
    _{2}=54.86 \%$). The first column illustrates the input surface point cloud,
    the second column presents the ground truth from finite element analysis,
    the third column shows the GINOT prediction, and the last column highlights the
    absolute error between the prediction and the ground truth.}
    \label{fig:jeb_plot}
  \end{figure}

  The dataset includes CAD models, volume meshes in VTK format, and solution
  fields on fine quadratic tetrahedral elements. Instead of extracting surface nodes
  from the volume mesh, we extract the boundary point cloud directly from the
  CAD model vertices for each JEB sample using FreeCAD with PythonAPI, as illustrated
  in \cref{fig:jeb_data}(a). To reduce computational costs, the volume mesh is
  downgraded from 10-node tetrahedrons to 4-node tetrahedrons by removing the
  middle nodes of each element's edge. The nodes of these linear tetrahedrons are
  then used as query points in the solution decoder of GINOT, as shown in
  \cref{fig:jeb_data}(a). \cref{fig:jeb_data}(b) illustrates the number of boundary
  points in the cloud and the volume mesh nodes for the 2,138 JEB samples. The
  boundary point cloud contains between 2,253 and 10,388 points, while the volume
  mesh nodes range from 18,085 to 55,009.

  \cref{fig:jeb_loss} illustrates the training process of GINOT on the JEB
  dataset. The mean squared error (MSE) loss for both the training and validation
  datasets decreases steadily during training. However, the testing loss remains
  significantly higher than the training loss, suggesting overfitting. %
  {\color{\HighlightColor} %
  This behavior is attributed to GINOT's high representational capacity—its geometry-aware encoder and transformer decoder can closely fit intricate training patterns. The JEB dataset is comparatively small (2,138 samples) yet contains highly complex geometries. This combination of high model capacity and limited, heterogeneous data increases susceptibility to overfitting. %
  }%

  After training, we evaluate GINOT's performance on both the training and testing
  samples. For the training samples, the mean $L_{2}$ relative error is 18.77\%
  with a standard deviation of 4.13\%, while for the testing samples, the mean $L
  _{2}$ relative error is 30.77\% with a standard deviation of 8.86\%.
  \cref{fig:jeb_plot} illustrates the Mises stress solutions for the testing
  samples predicted by GINOT, showcasing the best, median, and worst cases in terms
  of $L_{2}$ relative error. The best case achieves $L_{2}=17.0\%$, while the worst
  case has $L_{2}=54.0\%$.

  {\color{\HighlightColor}%
  To assess efficiency relative to SDF-based neural operators, the signed distance function was computed on a $128^{3}$ grid for first 3D test geometry (39k surface mesh triangles). The SDF evaluation required about $30\,\text{s}$ on an A100 GPU, whereas GINOT's sampling and grouping step took about $9\,\text{ms}$. This speed advantage shows that the proposed sampling and grouping strategy is substantially more efficient than explicit SDF construction for complex 3D geometries. %
  }%

  \subsection{Ablation study}

  Finally, we conducted an ablation study to evaluate the impact of different configurations
  of our model.

  The geometry encoder in our model includes cross-attention blocks followed by self-attention
  blocks. To assess the importance of these components, we performed experiments
  on the 2D PUC dataset and the 3D lug dataset. The results, summarized in \cref{tab:ablation_attention},
  are presented in terms of $L_{2}$ relative errors.

  The study reveals that the cross-attention block is crucial, as its removal significantly
  increases the $L_{2}$ error. This is because the cross-attention block
  effectively encodes global geometric features from the boundary point cloud. On
  the other hand, the self-attention block's necessity varies: its removal increases
  the $L_{2}$ error for the PUC dataset but decreases it for the lug dataset.
  This discrepancy is likely attributed to the higher complexity and randomness
  of the 2D PUC dataset compared to the parametric 3D lug dataset, where the self-attention
  block aids in learning better geometric representations for the more complex dataset.

  \begin{table}[h]
    \centering
    \caption{Ablation study results showing the $L_{2}$ relative errors for
    different attention block configurations.}
    \begin{tabular}{ccc}
      \hline
                                          & 2D PUC [$\%$] & 3D LUG [$\%$] \\
      \hline
      Full model                          & 7.81          & 0.40          \\
      Without cross-attention block       & 12.50         & 0.42          \\
      Without self-attention block        & 10.77         & 0.384         \\
      Without both cross- and self-blocks & 21.84         & 0.389         \\
      \hline
    \end{tabular}
    \label{tab:ablation_attention}
  \end{table}

  \begin{table}[h]
    \centering
    \caption{%
    {\color{\HighlightColor}%
    Ablation study results showing the $L_{2}$ relative errors (\%) for different sampling and grouping configurations: number of sampled centroids $N_{s}$, points per group $N_{p}$, and grouping radius $r$. "--" indicates that the configuration was not tested.%
    }%
    }
    \begin{tabular}{ccc|ccc|ccccc}
      \hline
      $N_{s}$ & PUC  & Elasticity & $N_{p}$ & PUC  & Elasticity & $r$ & PUC  & Elasticity & LUG  & JEB   \\
      \hline
      8       & --   & 1.09       & 8       & 7.81 & 1.27       & 0.1 & 7.61 & 1.25       & 0.41 & 30.77 \\
      16      & 8.99 & 1.03       & 18      & 8.33 & 1.33       & 0.2 & 7.81 & 1.33       & 0.43 & 30.84 \\
      32      & 8.27 & 1.16       & 32      & 7.46 & 1.36       & 0.4 & 8.98 & 1.30       & --   & --    \\
      64      & 7.97 & 1.33       & 64      & 7.40 & 1.25       & 0.5 & 9.23 & 1.53       & 0.40 & 31.70 \\
      96      & --   & 1.49       &         &      &            &     &      &            &      &       \\
      128     & 7.81 & --         &         &      &            &     &      &            &      &       \\
      \hline
    \end{tabular}
    \label{tab:ablation_attention_SG}
  \end{table}

  \begin{table}[h]
    \centering
    \caption{{\color{\HighlightColor}%
    Ablation on PC density for dataset of PUC. Reported are $L_{2}$ relative errors (\%) when evaluating with the same density as training or with the full (100\%) point cloud.}}
    \begin{tabular}{c|cc}
      \hline
      \makecell{Training PC density\\(\% of original)} & \makecell{$L_{2}$ when eval.\\with same density} & \makecell{$L_{2}$ when eval.\\with 100\% PC} \\
      \hline
      100\%                                            & 7.94                                             & 7.94                                         \\
      80\%                                             & 8.49                                             & 8.99                                         \\
      60\%                                             & 9.65                                             & 9.44                                         \\
      40\%                                             & 11.46                                            & 10.22                                        \\
      \hline
    \end{tabular}
    \label{tab:pc_density_ablation}
  \end{table}

  {\color{\HighlightColor}%
  We further examine the impact of the sampling and grouping hyperparameters in Table \ref{tab:ablation_attention_SG}. We vary the number of sampled centroids ($N_{s}$), the points per group ($N_{p}$), the grouping radius ($r$), and report the relative $L_{2}$ error on the PUC, Elasticity, LUG, and JEB datasets. For Elasticity (105 boundary points), changing percentage of sampling points ($N_{s}/105$) from 7.6\% to 30\% yields similar accuracy, while increasing it to 61-91\% degrades performance, likely because infomation become overly global. For PUC (up to 484 boundary points), increasing $N_{s}/484$ from 3.3\% to 26.4\% steadily improves accuracy, with diminishing returns beyond roughly 20-26\%. A practical choice is $N_{s}$ at about 15-30\% of the boundary points. The effect of $N_{p}$ is relatively minor on both PUC and Elasticity within the tested range. The grouping radius $r$ controls neighborhood size: smaller radii ($r=0.1$-$0.2$) generally perform best, whereas larger radii ($r= 0.4$-$0 .5$) tend to blend unrelated regions and may slightly hurt accuracy. We therefore recommend $r\in[ 0.1,0.2]$ together with a moderate $N_{s}$ ratio.

  We then studied the effect of point-cloud (PC) density. Performance varies nonlinearly with density. As shown in \cref{tab:pc_density_ablation}, reducing the PC density from 100\% to 80\% leaves the relative $L_{2}$ error essentially unchanged, indicating robustness to moderate downsampling. When the density drops to 40\%, errors increase markedly due to the loss of fine geometric detail needed to capture local structure. Thus, modest downsampling can reduce compute with minimal accuracy loss, but aggressive sparsification should be avoided. Evaluating with denser PCs than those used in training only slightly affects accuracy when the training PC density is high (e.g., $\ge 60\%$), but yields accuracy improvement when the training PC density is low (e.g., 40\%). %
  These results confirm that reducing point-cloud density, which removes geometric detail, degrades accuracy. %
  }%

  \section{Conclusions}

  In this work, we introduced the Geometry-Informed Neural Operator Transformer
  (GINOT), a novel framework for efficient and accurate solution prediction on arbitrary
  geometries represented by unordered, non-uniform, and variable-sized surface point
  clouds, without the need for additional geometric features such as signed
  distance functions (SDFs). By utilizing transformer-based attention mechanisms
  and a geometry-aware encoding strategy, GINOT effectively captures geometric
  representations from boundary point clouds and maps them to solution fields at
  arbitrary query points. Unlike existing neural operator methods constrained by
  structured meshes, fixed query points, or computationally intensive geometric representations,
  GINOT offers a flexible and scalable approach for solving complex geometry
  problems. To enhance its functionality, we have integrated additional encoders
  to process problem-specific inputs, such as material properties and boundary conditions.

  Extensive experiments on diverse datasets have demonstrated GINOT's good accuracy
  and generalization capabilities across various geometries. These results
  establish GINOT as a powerful surrogate model for computational physics
  involving complex domains, providing significant computational speedups while maintaining
  high predictive accuracy. Future work will focus on extending GINOT to tackle
  multi-physics problems, further enhancing its utility in scientific computing and
  engineering applications.

  \section{Data availability}
  The dataset is available on Zenodo \citep{liu2025geometry} and the trained
  models are available at the GitHub repository
  \url{https://github.com/QibangLiu/GINOT}

  \section{Code availability}
  The codes for training and inference are available at the GitHub repository at
  \url{https://github.com/QibangLiu/GINOT}

  \section{Acknowledgements}
  The authors would like to thank the National Center for Supercomputing
  Applications (NCSA) at the University of Illinois, and particularly its Research
  Computing Directorate, Industry Program, and Center for Artificial Intelligence
  Innovation (CAII) for support and hardware resources. This research used both
  the DeltaAI advanced computing and data resource, which is supported by the National
  Science Foundation (award OAC 2320345) and the State of Illinois, and the
  Delta advanced computing and data resource which is supported by the National
  Science Foundation (award OAC 2005572) and the State of Illinois. Delta and DeltaAI
  are joint efforts of the University of Illinois Urbana-Champaign and its National
  Center for Supercomputing Applications. This work used Delta/DeltaAI at NCSA
  through allocations MAT240113 and CHE230009 from the Advanced
  Cyberinfrastructure Coordination Ecosystem: Services \& Support (ACCESS) program,
  which is supported by U.S. National Science Foundation grants \#2138259, \#2138286,
  \#2138307, \#2137603, and \#2138296.

  \section{Author contributions statement}
  \textbf{Q. Liu} conceptualized the study, developed the methodology,
  implemented the models, prepared the datasets, conducted the analysis, and drafted,
  reviewed, and edited the manuscript. \textbf{W. Zhong} contributed to running the
  baseline models, manuscript review and editing. \textbf{H. Meidani}
  contributed to conceptualization, results discussion, and manuscript review
  and editing. \textbf{D. Abueidda} contributed to results discussion and
  manuscript review and editing. \textbf{S. Koric} prepared the bracket lug
  dataset, contributed to conceptualization, results discussion, and reviewed and
  edited the manuscript. \textbf{P. Geubelle} contributed to the supervision,
  methodology, results discussion, and manuscript review and editing.

  \section{Competing interests statement}
  The authors declare no competing financial or non-financial interests.

  \bibliographystyle{elsarticle-num-names}


  \bibliography{references}

\end{document}